\documentclass{article}

\usepackage[final]{nips_2017}

\usepackage[utf8]{inputenc} 
\usepackage[T1]{fontenc}    
\usepackage{hyperref}       
\usepackage{url}            
\usepackage{booktabs}       
\usepackage{amsfonts}       
\usepackage{nicefrac}       
\usepackage{microtype}      
\usepackage{graphicx}

\title{Label scarcity in biomedicine: Data-rich latent factor discovery
enhances phenotype prediction}

\author{
  Marc-Andre~Schulz\\
  Department of Psychiatry and Psychotherapy\\
  RWTH Aachen University\\
  Aachen, Germany \\
  \And
  Bertrand Thirion,
  Alexandre Gramfort,
  Ga\"el Varoquaux\\
  INRIA, Parietal team, Saclay, France\\
  CEA, Neurospin, Gif-sur-Yvette, France\\
  \And
  Danilo Bzdok \\
  Department of Psychiatry and Psychotherapy\\
  RWTH Aachen University\\
  Aachen, Germany \\
}

\begin{document}

\maketitle

\begin{abstract}
  High-quality data accumulation is now becoming ubiquitous in the health domain.
  There is increasing opportunity to exploit rich data from normal subjects
  to improve supervised estimators in specific diseases with notorious data scarcity.
  We demonstrate that low-dimensional embedding spaces can be derived
  from the UK Biobank population dataset and used
  to enhance data-scarce prediction of health indicators, lifestyle and
  demographic characteristics. Phenotype predictions facilitated by
  Variational Autoencoder manifolds typically scaled better with
  increasing unlabeled data than dimensionality reduction
  by PCA or Isomap.
  Performances gains from semisupervison approaches will probably become
  an important ingredient for various medical data science applications.
\end{abstract}

\section{Introduction}

Massive data accumulation, an existing trend in industry and government, is now gaining traction in medicine. Among the recently emerged extensive high-quality biomedical datasets, the UK Biobank\footnote{www.ukbiobank.ac.uk} (UKBB) is probably the most prominent: A longitudinal prospective population study with extensive profiling in 500,000 subjects.
Growing data richness opens the door to deploying machine learning techniques to gain new insight into disease. The need for new data-driven methods is particularly urgent in medical specialties where traditional diagnostic criteria present a poor reflection of the underlying biological dysfunction, such as in mental health (Hyman 2007). In the example of psychiatry, latent factor discovery on large scale imaging data may allow to go beyond current diagnostic catalogues by revealing and exploiting objectively measurable biomarkers, potentially enabling early diagnosis and individualized, biologically informed treatment and prognosis.
This agenda has emerged as \textit{precision medicine}.

While emerging population datasets provide a wealth of data from \textit{healthy} subjects, the amount of high-quality descriptions from \textit{diseased} subjects remains a major limiting factor to deploy advanced prediction models in most medical specialties. The small numbers of patients and matched controls limits the use of modern, data-hungry machine learning algorithms. Therefore, precision medicine is facing a major question: Can we exploit the already abundant \textit{unlabeled} data from l-purpose datasets of the normal population to inform analysis in the many \textit{label-scarce} medical prediction problems (healthy vs. disease)?
\textit{Semisupervised} machine learning provides a possible solution. These algorithms lly combine unsupervised structure discovery and supervised prediction, attempting to learn a representation of the data with the help of unlabeled samples that is useful for the prediction task at hand (labeled samples). Recently, powerful deep semisupervised estimators have been proposed (Kingma et al. 2014; Rasmus et al. 2015; Miyato et al. 2015) that could in the future be applied to label-scarce biomedical data.

The present study benchmarks performance improvements of combined unsupervised-supervised learning strategies on small-data prediction problems in biomedicine obtained from latent \textit{endo-phenotypes} hidden in reference datasets. Given widespread label scarcity in medical learning problems, this important application field will particular benefit from novel combined approaches for structure discovery and large-scale prediction.

\section{Methods}

Three reference datasets were used for the present analyses: Mnist and Zalandos Fashion\footnote{www.github.com/zalandoresearch/fashion-mnist} reference datasets (both 784 pixels per grayscale picture) and brain imaging data from UKBB: Three major imaging modalities (\textit{T1-MRI} - brain volume; \textit{rfMRI} - neural activity fluctuations in the resting brain; \textit{dMRI} -  fiber tract anatomy)  were considered. In each case, precomputed neurobiologically meaningful features (\textit{imaging derived phenotypes, IDPs}) were used ($p = 164; 210; 432$) instead of raw pixels ($p = 10^7; 10^8; 10^5$) to keep feature dimensionality in the same order of magnitude as Mnist/Fashion. Prediction targets sex, age, past tobacco smoking, work satisfaction, household income, and number of people in household (2, 5, 4, 5, 5, 5 classes respectively) were selected to represent a range of demographic, lifestyle and health-related phenotypes.

For these prediction targets and imaging modalities there were 7500 subjects available in the UKBB. Data was randomly split into a 7000 sample train and a 500 sample test set (25 repetitions for uncertainty estimation). Latent factor embeddings were created for varying proportions of the train set. Three representative embedding algorithms were used: the classical linear dimensionality reduction method \textit{Principal Component Analysis} (PCA, Pearson 1901), the more modern non-linear embedding \textit{Isomap} (Tenenbaum, de Silva, and Langford 2000; 5 nearest neighbors) and the recent  neural network \textit{Variational Autoencoders} (VAE, Kingma and Welling 2014; 1 hidden layer â 200 units, Relu activation, Adam) were used for factor discovery.
In a series of sample complexity analyses,
the generated embeddings were
used for fitting classification models with
varying numbers of labeled training samples. $\ell_2$-regularized Logistic Regression and Random Forest classifiers were used for classification. For each prediction problem, training sets of 100-1000 labeled samples were selected. Based on these and a growing number (100-7000) of additional unlabeled samples, embedding spaces were fitted.  These were subsequently used  as input for the classifiers.

\section{Results}

\begin{figure}
  \centering
   \includegraphics[width=1\textwidth]{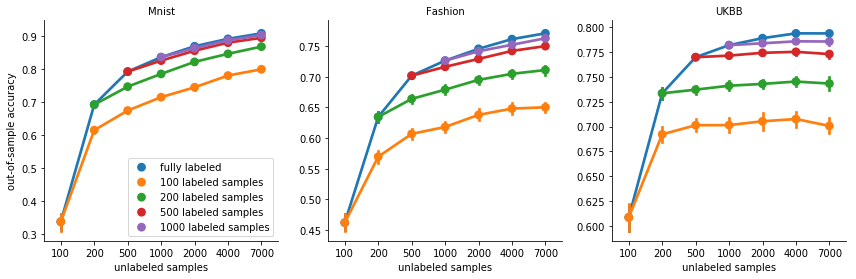}
  \caption{\textit{Discovering latent factors from data richness enhances prediction accuracy in label scarcity.} In reference datasets in machine learning (Mnist, Fashion) and medicine (UKBB), Isomap conducted non-linear dimensionality reduction, 
  and the ensuing parsimonous embeddings of
  high-dimensional data were used for out-of-sample prediction of
  numbers (left), clothing types (middle), or subject sex (right).
  Learning such low-dimensional embeddings on increasing amounts of unlabeled data considerably boosted classification in the fixed test samples.
  Abundance of unlabeled samples was
  more beneficial in
  simpler prediction problems (Mnist, Fashion) than for biomedical phenotypes.}
\end{figure}

In a first step, we examined the complexity of the reference datasets Mnist, Fashion, and UKBB Imaging. Raw pixels (T1-MRI IDPs respectively) were used for factor discovery. Isomap was used on 100-7000 unlabeled samples to embed data into a 50 dimensional latent space. A 7000 sample Isomap embedding into a 50 dimensional latent space with subsequent Logistic Regression bridges $(89.89\pm 0.78)\%$ of the accuracy gap (“\textit{semisupervision effect}”)  between 100 and 7000 labeled samples. In the Fashion dataset we observe $(60.89\pm1.38)\%$, in UKBB T1-Sex prediction $(49.52\pm3.21)\%$.
The semisupervision effect decreases from Mnist to the more complex Fashion dataset and from Fashion to UKBB indicating that reliable structure discovery and prediction from biomedical data may be intrinsically more challenging. 

\begin{figure}
  \centering
   \includegraphics[width=1\textwidth]{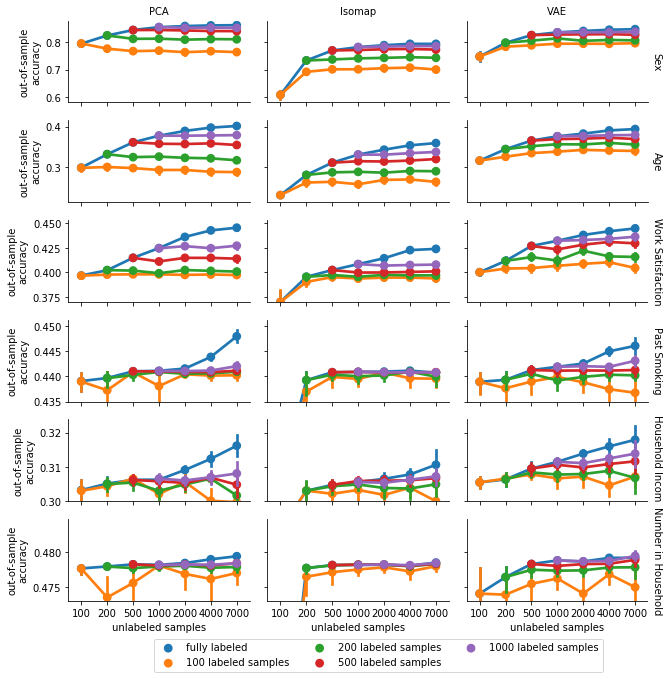}
  \caption{\textit{Semisupervised performance gains with variational autoencoders.} In brain volume data (T1) from 7500 subjects, dimensionality reduction enforcing decorrelation (PCA) was benchmarked against compression into
  an embedding space allowing for non-linear effects (Isomap)
  and higher-order relations between the input dimensions (VAE).
  More sophisticated embeddings (VAE)
  more effectively translated
  more unlabeled samples into accuracy gains.
  For an identical embedding, the extent of accuracy improvement depended on the prediction target. The x-axis indicates the number of unlabeled samples used to extract the embedding. In each learning problem, algorithms are fitted on
  a series of
  training sets with 100 to 1,000 labeled examples
  and evaluted on fixed test samples.}
\end{figure}

In an second step, we characterize the sample complexity of prediction targets (sex, age, income household size, work satisfaction) from brain morphology involving distinct decomposition methods: PCA, Isomap, and VAE.
PCA shows no semisupervised gains. In some cases (sex, age) more unlabeled data even harms prediction performance. Isomap does indeed show show semisupervised gains, but consistently performs under the PCA baseline. VAE yields superior prediction performance for all targets, even without extra unlabeled data. It shows significant semisupervised gains (across target weighed average: $(30.24\pm2.85)\%$; $(17.43\pm3.53)\%$; $(11.43\pm5.53)\%$; $(23.58\pm6.48)\%$ for 100, 200, 500, 1000 labeled samples respectively). Higher-order structure readily captured by VAE globally profits more from increased unlabeled data for phenotype prediction.

\begin{figure}
  \centering
   \includegraphics[width=0.68\textwidth]{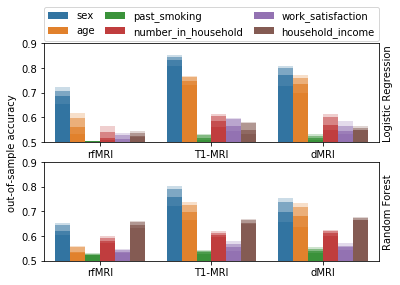}
  \caption{\textit{UKBB phenotypes can profit from nonlinear estimators.}
  In neural activity fluctuations in the resting brain (rfMRI), brain volume
  (T1-MRI), and fiber tract anatomy (dMRI),
  unsupervised VAE embeddings were derived
  before phenotype prediction via linear (Logistic Regression, upper row)
  and non-linear (Random Forest, lower row) prediction.
  Demographic target variables (household income/size)
  benefitted most from classifiers admitting non-linear structure.
  Transparency of the bars indicates
  the number of labeled samples (100 vs. 200 vs. 500 vs. 1000).}
\end{figure}

In the final step, we used the low dimensional latent space to compare prediction performance across targets and imaging modalities with both a linear (Logistic Regression) and non-linear (Random Forest) classifier. For comparability over different numbers of classes we report binary “above/below average” prediction accuracy. 
Prediction performance across targets was best in T1-MRI with a weighted average cross-target accuracy $(64.80\pm0.25)\%$ (rfMRI: $(53.46\pm0.25)\%$; dMRI: $(58.49\pm0.24)\%$).
Across targets, Logistic Regression and Random Forests show differing performance profiles. The linear classifier performs better on simple targets (sex, age), whereas the gain of non-linear classification is especially observed for challenging prediction targets (income, household size).

\section{Discussion}

The present investigations demonstrate that emerging general-purpose datasets
in health can be exploited for challenging phenotype predictions
plagued by data scarcity.
Extracting sophisticated latent factors from high-dimensional biomedical data
can conveniently tackle two distinct problems at once.
First, for many biomedical phenotypes it is currently unknown which type of
genetic detail, body scans, demographic information, or routine performance
tests will turn out to be most useful for reliable prediction.
State-of-the-art dimensionality reduction techniques will probably
be applied to hetereogenous data from diverging sources 
to re-express high-dimensional data in quintessential sources of inter-individual
variation.
Second, there is uncertainty in medical specialties whether the established
disease categories are a faithful description of the underlying biological
disturbance, such as in psychiatric diagnoses and neuroscience.
Latent factors of population variation can be derived from accumulating
biomedical data repositories to automatically extract dimensions that
effectively describe single individuals transcending binary
health and disease classification.

\newpage

\section*{References}

\small
Hyman, Steven E. 2007. “Can Neuroscience Be Integrated into the DSM-V?” Nature Reviews. Neuroscience 8 (9): 725–32.

Kingma, Diederik P., Shakir Mohamed, Danilo Jimenez Rezende, and Max Welling. 2014. “Semi-Supervised Learning with Deep Generative Models.” In Advances in Neural Information Processing Systems 27: 3581–89.

Kingma, Diederik P., and Max Welling. 2013. “Auto-Encoding Variational Bayes.” Auto-encoding variational Bayes. In Proceedings of the International Conference on Learning Representations (ICLR).

Miyato, Takeru, Shin-Ichi Maeda, Masanori Koyama, Ken Nakae, and Shin Ishii. 2015. “Distributional Smoothing with Virtual Adversarial Training.” arXiv [stat.ML]. arXiv. http://arxiv.org/abs/1507.00677.

Pearson, K. 1901. “On Lines and Planes of Closest Fit to Systems of Points in Space.” Philosophical Magazine 2 (11): 559–72.

Rasmus, Antti, Mathias Berglund, Mikko Honkala, Harri Valpola, and Tapani Raiko. 2015. “Semi-Supervised Learning with Ladder Networks.” In Advances in Neural Information Processing Systems 28: 3546–54.

Tenenbaum, J. B., V. de Silva, and J. C. Langford. 2000. “A Global Geometric Framework for Nonlinear Dimensionality Reduction.” Science 290 (5500): 2319–23.

\end{document}